\newcommand{\CLASSINPUTinnersidemargin}{0.75in} % inner side margin
\newcommand{\CLASSINPUToutersidemargin}{0.75in} % outer side margin
\newcommand{\CLASSINPUTtoptextmargin}{0.75in} % top text margin
\newcommand{\CLASSINPUTbottomtextmargin}{0.75in}% bottom text margin

\documentclass[letterpaper, 10pt, conference]{IEEEtran}
\IEEEoverridecommandlockouts
\usepackage{amsmath, amssymb, amsfonts, amsthm, mathtools, bm}
\usepackage{graphicx}
\usepackage{tabularx, threeparttable, booktabs}
\usepackage{multirow, multicol}
\ifCLASSOPTIONcompsoc
\usepackage[caption=false,font=normalsize,labelfont=sf,textfont=sf]{subfig}
\else
\usepackage[caption=false,font=footnotesize]{subfig}
\fi

\usepackage[linesnumbered, lined, boxed, ruled]{algorithm2e}

\usepackage{cite}
\usepackage{textcomp}
\usepackage[dvipsnames]{xcolor}
\usepackage{url}
\usepackage{lipsum}

\newcommand{\prl}[1]{\mathopen{}\left(#1\right)\mathclose{}}

\graphicspath{{figures/}{figures/highres_cropped/}}

\begin{document}

\title{\vspace{6mm}RTGNN: A Novel Approach to Model \\Stochastic Traffic Dynamics
\thanks{Ke Sun and Vijay Kumar are with GRASP Lab,
        University of Pennsylvania, Philadelphia, PA 19104, USA,
        {\tt\footnotesize\{sunke, kumar\}@seas.upenn.edu.}
        Stephen Chaves and Paul Martin are with Qualcomm Technologies Inc.,
        Philadelphia, PA 19146, USA,
        {\tt\footnotesize\{schaves, pdmartin\}@qti.qualcomm.com.}}
        %We gratefully acknowledge the support of Qualcomm Research who sponsored this work.}%
%\thanks{Stephen Chaves and Paul Martin are with Qualcomm Technologies Inc.,
%        Philadelphia, PA 19146, USA,
%        {\tt\small\{schaves, pdmartin\}@qti.qualcomm.com.}}% <-this % stops a space
}
\author{Ke~Sun, Stephen~Chaves, Paul~Martin, and~Vijay~Kumar}

%\author{
%\IEEEauthorblockN{1\textsuperscript{st} Given Name Surname}
%\IEEEauthorblockA{\textit{dept. name of organization (of Aff.)} \\
%\textit{name of organization (of Aff.)}\\
%City, Country \\
%email address or ORCID}
%\and
%\IEEEauthorblockN{2\textsuperscript{nd} Given Name Surname}
%\IEEEauthorblockA{\textit{dept. name of organization (of Aff.)} \\
%\textit{name of organization (of Aff.)}\\
%City, Country \\
%email address or ORCID}
%\and
%\IEEEauthorblockN{3\textsuperscript{rd} Given Name Surname}
%\IEEEauthorblockA{\textit{dept. name of organization (of Aff.)} \\
%\textit{name of organization (of Aff.)}\\
%City, Country \\
%email address or ORCID}
%\and
%\IEEEauthorblockN{4\textsuperscript{th} Given Name Surname}
%\IEEEauthorblockA{\textit{dept. name of organization (of Aff.)} \\
%\textit{name of organization (of Aff.)}\\
%City, Country \\
%email address or ORCID}
%\and
%\IEEEauthorblockN{5\textsuperscript{th} Given Name Surname}
%\IEEEauthorblockA{\textit{dept. name of organization (of Aff.)} \\
%\textit{name of organization (of Aff.)}\\
%City, Country \\
%email address or ORCID}
%\and
%\IEEEauthorblockN{6\textsuperscript{th} Given Name Surname}
%\IEEEauthorblockA{\textit{dept. name of organization (of Aff.)} \\
%\textit{name of organization (of Aff.)}\\
%City, Country \\
%email address or ORCID}
%}

\maketitle

\begin{abstract}
  Modeling stochastic traffic dynamics is critical to developing self-driving cars.
  Because it is difficult to develop first principle models of cars driven by humans, there is great potential for using data driven approaches in developing traffic dynamical models.
  While there is extensive literature on this subject, previous works mainly address the prediction accuracy of data-driven models.
  Moreover, it is often difficult to apply these models to common planning frameworks since they fail to meet the assumptions therein.
  In this work, we propose a new stochastic traffic model, Recurrent Traffic Graph Neural Network (RTGNN), by enforcing additional structures on the model so that the proposed model can be seamlessly integrated with existing motion planning algorithms.
  RTGNN is a Markovian model and is able to infer future traffic states conditioned on the motion of the ego vehicle.
  Specifically, RTGNN uses a definition of the traffic state that includes the state of all players in a local region and is therefore able to make joint predictions for all agents of interest.
  Meanwhile, we explicitly model the hidden states of agents, ``intentions,'' as part of the traffic state to reflect the inherent partial observability of traffic dynamics.
  The above mentioned properties are critical for integrating RTGNN with motion planning algorithms coupling prediction and decision making.
  Despite the additional structures, we show that RTGNN is able to achieve state-of-the-art accuracy through comparisons with other similar works.
\end{abstract}

%\begin{IEEEkeywords}
%\end{IEEEkeywords}

\section{Introduction}
\label{sec: introduction}

% Why stochastic traffic dynamics modeling is an important problem.
In order to safely navigate an autonomous vehicle (ego), it is of crucial importance to understand the behavior of other players (agents) in the traffic.
Addressing this problem requires modeling the evoluation of traffic, \textit{i.e.} traffic dynamics.
Especially, a stochastic model is necessary considering the uncertain nature of traffic dynamics.

Simple parametric models, such as IDM~\cite{Treiber-Springer-2013} and~MOBIL~\cite{Kesting-TRR-1999}, may serve this purpose.
Works like~\cite{Sun-IROS-2020, Zhang-ICRA-2020} use such models for traffic dynamics and plan motions for the ego vehicle therein.
However, these models are designed for studying the macro traffic behavior.
Predictions of the models might not agree with the local behavior of agents in real traffic scenarios.
Limitations of handcrafted parametric models motivate data-driven approaches in predicting agent behavior, which have recently become possible thanks to the large scale datasets like ArgoVerse~\cite{Chang-CVPR-2019} and nuScenes~\cite{Caesar-CVPR-2020}.

% Limitations of the exisiting works.
Most existing data driven models for traffic dynamics focus on improving the prediction accuracy, \textit{i.e.} reducing the difference between the predicted trajectories and the ground truth for agents.
However, it is also important that the model satisfies assumptions of common motion planning algorithms, considering the traffic model is eventually to be used in planning.
Most planning algorithms~\cite{Lavalle-Cambridge-2006} assume the stochastic dynamical model is in the form of,
\begin{equation}
  \label{eq: general dynamical models}
  \bm{x}_{t+1} = f(\bm{x}_t, \bm{u}_t, \bm{n}_t),
\end{equation}
where $\bm{x}$ and $\bm{u}$ are state and control of the system, while $\bm{n}$ is a random variable modeling the motion noise.
In the context of traffic dynamics, the motion model in~\eqref{eq: general dynamical models} imposes the following assumptions on the model structure and state representation.
In terms of structure, the traffic model should be able to make predictions for the next time step based on the current traffic state and immediate motion of the ego vehicle.
The former is also known as the Markovian property of a dynamical model.
In terms of state representation, the traffic state should consist of the states of all agents in a local neighborhood, which are necessary for decision making at the level of motion planning.
It is also important for the state representation to reflect the partially observable characteristic of traffic.
Internal states of agents, often known as ``intentions'', are not observable.
Modeling the internal states provides the possibility to estimate them online and reduce the uncertainty of future predictions.
Few existing data driven traffic models satisfy all above mentioned assumptions, making them hard to be tightly integrated into motion planning algorithms.

\begin{figure}[t]
  \centering
  \includegraphics[width=0.48\columnwidth]{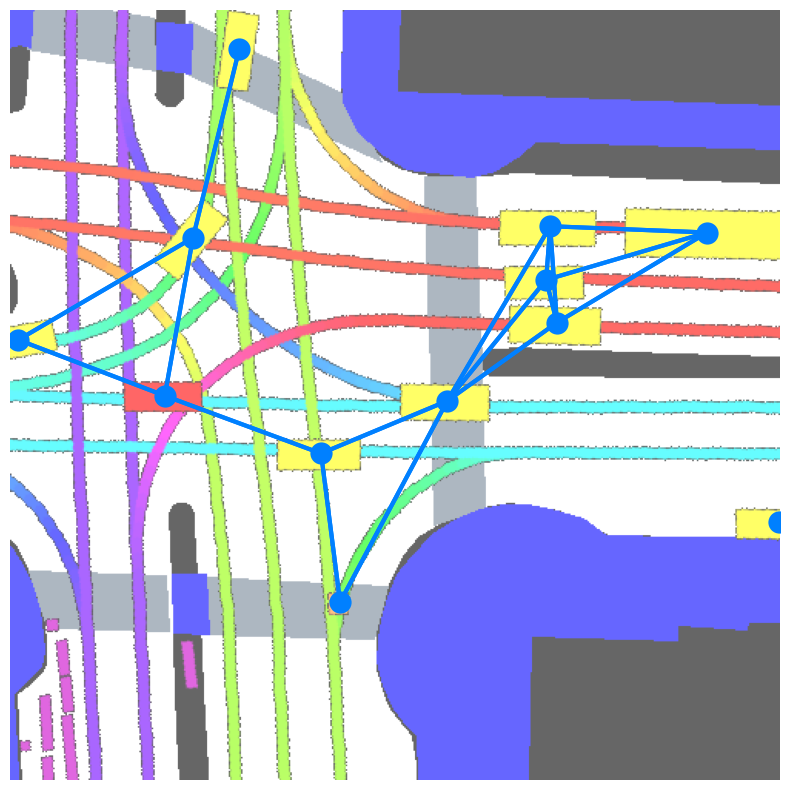}
  \includegraphics[width=0.48\columnwidth]{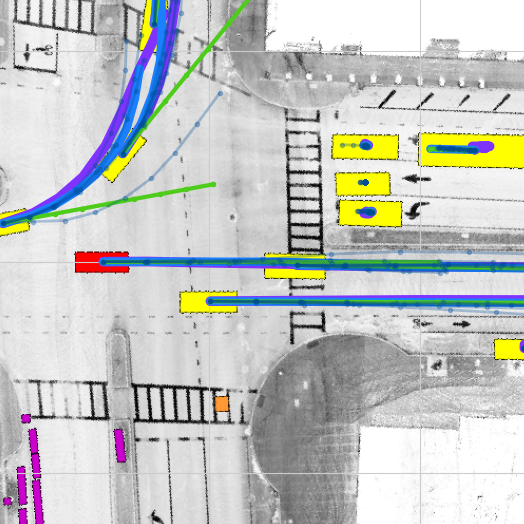}
  \caption{RTGNN (left) builds directed graphs to represent the local traffic and (right) makes joint predictions for all agents in the scene conditioned on the current traffic state and the immediate ego motion. Red and yellow rectangles represent the ego and agent vehicles, while orange squares represent pedestrians. (Right) purple lines are the ground truth. Green lines are predictions of a constant velocity model. Blue and gray lines are the maximum likelihood and five sampled predictions of RTGNN.}
  \label{fig: annotation narrative}
\end{figure}

% Contributions of this work.
\textbf{Contributions}: In this work, we propose Recurrent Traffic Graph Neural Network (RTGNN) to model the stochastic traffic dynamics.
The proposed traffic model is in the form of~\eqref{eq: general dynamical models}.
Structure-wise, the model is Markovian and is able to infer agent behavior conditioned on the motion of the ego vehicle.
The state of the proposed traffic model consists of information of both ego and other agents.
Therefore, the model is able to produce joint predictions for all agents within a local region.
Meanwhile, we explicitly model the hidden states, ``intentions'', of agents to reflect the partial observable property of traffic dynamics.

With the above properties, RTGNN could be seamlessly integrated into common motion planning algorithms coupling predictions and decision making.
In terms of learning, the additional structure and explicit hidden state representation impose regulations on the model.
Nevertheless, we show that RTGNN is still able to achieve more accurate predictions through comparisons with other similar works.

\section{Related Works}
\label{sec: related works}

Traffic models in related works are often non-Markovian~\cite{Chai-CoRL-2019, Cui-ICRA-2019, Phan-CVPR-2020, Rhinehart-ICCV-2019}, which predict the future in the next few seconds based on the past agent trajectories of a short duration.
The non-Markovian structure allows more flexibility for the network to exploit data over multiple steps and could therefore potentially improve prediction accuracy.
However, it may be hard to apply efficient planning algorithms implementing concepts like dynamic programming with such models because of the lack of Markovian structure.
%However, models with such properties can no longer be tightly integrated into motion planning algorithms.
Recurrent models~\cite{Sanchez-Gonzalez-PMLR-2018} are suitable for this purpose, which take the network output as the input for the next time step.
However, recurrent models are known to be subject to compounding errors~\cite{Ross-PMLR-2011}.
Although not fully resolved, the issue can be alleviated with scheduled sampling as shown by Bengio~\cite{Bengio-NIPS-2015}.

Another important aspect of the traffic model is to reflect the interactions between the ego and agents, \textit{i.e.} make predictions of the traffic conditioned on the motion of the ego vehicle.
The property is investigated by Rhinehart~\cite{Rhinehart-ICCV-2019}, Khandelwal~\cite{Khandelwal-ArXiv-2020}, Salzmann~\cite{Salzmann-ECCV-2021}, and Tolstaya~\cite{Tolstaya-ArXiv-2021}.
Rhinehart~\cite{Rhinehart-ICCV-2019} and Khandelwal~\cite{Khandelwal-ArXiv-2020} both infer agent motion conditioned on a local goal of the ego.
Simply conditioning on the goal might be insufficient, since various ego motions could lead to the same goal, while each of them could interact with the agents differently.
Instead, Salzmann~\cite{Salzmann-ECCV-2021} and Tolstaya~\cite{Tolstaya-ArXiv-2021} perform inference conditioned on the future trajectory of the ego.
It is straightforward to use both models for query, \textit{i.e.} predicting agent trajectories with a complete ego trajectory.
However, construting an ego trajectory incrementally would be hard to realize with such models.

A large body of prior works~\cite{Chai-CoRL-2019, Cui-ICRA-2019, Phan-CVPR-2020, Messaoud-arXiv-2020, Gao-CVPR-2020, Zhao-CoRL-2020} focus on predicting the future of a single agent.
However, a traffic scene may consist of various numbers of players, where predictions for all are required.
Although the previously mentioned algorithms can be applied to all agents sequentially, they are limited in capturing the complex interactions between agents.
As a result, the predictions may not be consistent, \textit{i.e.} predictions of different agents could lead to collisions.
To address the issue, more recent works~\cite{Salzmann-ECCV-2021, Casas-ICRA-2020, Casas-ECCV-2020, Liang-ECCV-2020, Zeng-ArXiv-2021} are able to make joint predictions for all agents of interest, which either explicitly or implicitly apply the idea of a GNN~\cite{Battaglia-ArXiv-2018}.

Works like~\cite{Casas-CoRL-2018, Rhinehart-ICCV-2019, Salzmann-ECCV-2021} try to model the partial observability of traffic dynamics by introducing latent variables into the network.
Rhinehart~\cite{Rhinehart-ICCV-2019} applies latent variables per agent per step in order to randomize the predictions.
Casas~\cite{Casas-CoRL-2018} and Slazmanni~\cite{Salzmann-ECCV-2021} introduce discrete latent variables intended to affect the high-level behavior of agent predictions.
In the first case~\cite{Rhinehart-ICCV-2019}, latent variables are \textit{i.i.d} and therefore independent of the traffic state.
In the second case~\cite{Casas-CoRL-2018, Salzmann-ECCV-2021}, the latent variables are internal to the network and generated as intermediate results.
In either case, it is not straightforward to include such latent variables as part of the traffic state.

\section{Modeling Traffic Dynamics}
\label{sec: modeling traffic dynamics}

For clarity, we begin by considering the traffic as an autonomous system, \textit{i.e.} the ego vehicle is not distinguished from other agents.
Discussions on adapting the model to perform inference conditioned on the ego motion are deferred to later in this section.

The dynamics of a single vehicle is modeled as a (dynamically extended) unicycle, which is of the form,
\begin{equation}
  \label{eq: unicycle dynamics for vehicles}
  \bm{x}_{t+1} = f_x(\bm{x}_t, \bm{u}_t).
\end{equation}
$\bm{x} = (x, y, \theta, v)^\top \in \mathcal{X}$ is the \textit{vehicle state} consisting of 2-D position $x$ and $y$, orientation $\theta$, and speed $v$.
The \textit{control input}, $\bm{u} = (a, \omega)^\top \in \mathcal{U}$, includes both linear acceleration, $a$, and angular velocity, $\omega$.
The explicit formulation of~\eqref{eq: unicycle dynamics for vehicles} could be found in~\cite{Lavalle-Cambridge-2006} and~\cite{Salzmann-ECCV-2021}.

Instead of having a continuous control space $\mathcal{U}$, we define $\mathcal{A}\subseteq\mathcal{U}$, a finite set consisting of predetermined \textit{motion primitives}, \textit{i.e.} $\mathcal{A} = \{(a_i, \omega_i), i=0, 1, \dots, M-1\}$.
Meanwhile, we define $\bm{q}\in\mathbb{R}^{|\mathcal{A}|}$ as a discrete probability distribution over $\mathcal{A}$.
In the following, we refer to $\bm{q}$ as the \textit{intention} of an agent, which determines the probability of taking each motion primitive at a single time step.
Combining the vehicle state and intention, an \textit{agent state} is defined as $\bm{y}^\top = (\bm{x}^\top, \bm{q}^\top)$.
In this work, we assume uncertainty of an agent only comes from its intention, or equivalently, the unknown motion primitive to be executed by the agent.

Presumably, the intention of an agent is affected by its neighbors.
We define $\bm{X}$, $\bm{Q}$, and $\bm{Y}$ as the aggregation of vehicle states, intentions, and agents states of all agents (including the ego) in local traffic.
We assume the dynamics of the intentions is of the form,
\begin{equation}
  \label{eq: traffic intention dynamics}
  \bm{Q}_{t+1} = f_q(\bm{Y}_t, \bm{m}_t) = f_q(\bm{X}_t, \bm{Q}_t, \bm{m}_t),
\end{equation}
where $\bm{m}$ represents the local semantic maps as shown in \figurename~\ref{fig: annotation narrative}, which include components like drivable areas, lane geometry, \textit{etc}.
The form of~\eqref{eq: traffic intention dynamics} implicitly assumes deterministic transition of intention.
However, it does not imply deterministic transition of agent motion, but rather deterministic transition of motion distribution.

To reduce the complexity of the traffic model, we only consider vehicles and assume a deterministic first-order motion model for pedestrians.
Equivalently, we could use~\eqref{eq: unicycle dynamics for vehicles} to model the motion of pedestrians but with a fixed intention concentrating all probability mass on the motion primitive $(0, 0)$, \textit{i.e.} the motion primitive with zero linear acceleration and angular velocity.
%Based on empirical observations from the dataset, the approximation provides sufficiently accurate predictions for pedestrians for a short horizon (around 5s).

\subsection{Intention Dynamics}
\label{subsec: modeling intention dynamics with a GNN}

\begin{figure}[t]
  \centering
  \includegraphics[width=0.48\columnwidth]{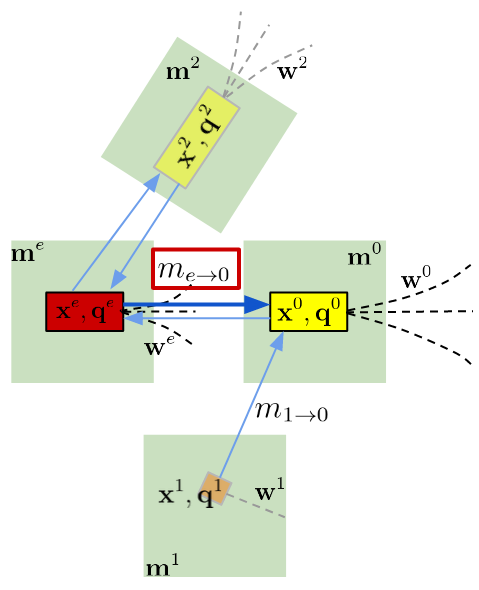}
  \includegraphics[width=0.48\columnwidth]{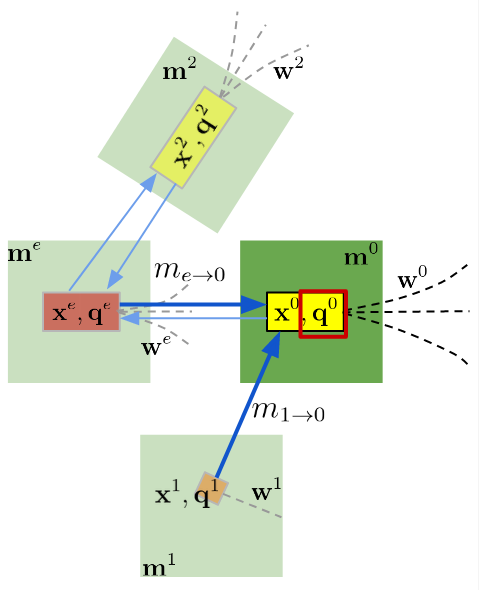}
  \caption{Schematic figures showing major steps in GNN for intention update. (Left) shows the message computation along an edge. (Right) shows the intention update at a node. Red, yellow, and orange boxes represent nodes of the ego vehicle, agent vehicles, and pedestrians. Curves next to each player are the future vehicle states under different motion primitives. Green patches represent the local agent-centric semantic maps. In the figures, the variables to be computed are marked with red boxes. Variables involved in the computation are highlighted while unrelated variables are grayed out. See Sec.~\ref{sec: modeling traffic dynamics} for the definitions of variables and Alg.~\ref{alg: modeling intention dynamics with a GNN} for the detailed steps.}
  \label{fig: example graph of local traffic}
\end{figure}

As mentioned earlier, $f_x$, modeled as a unicycle in this work, is relatively straightforward to specify.
However, $f_q$ involves human factors, making it hard to be specified in any simple parametric form.
Therefore, we structure $f_q$ as a deep neural network and learn the parameters through data.
In this work, we use a GNN for the intention dynamics, $f_q$.
The major motivation is that GNNs are able to handle graphs with various numbers of nodes and edges, as is the case in traffic scenes where there are multiple agents with complex interactions.

As shown in \figurename~\ref{fig: annotation narrative} and~\ref{fig: example graph of local traffic}, each node in the graph represents an agent in the local traffic, either a vehicle or a pedestrian, while edges are introduced based on spatial proximity. The feature of a node/agent includes vehicle state $\bm{x}^a = (x^a, y^a, \theta^a, v^a)$, future vehicle states at the next step under different motion primitives $\bm{w}^a$, intention $\bm{q}^a$, and a local agent-centric view of semantic map $\bm{m}^a$ similar to what is shown in \figurename~\ref{fig: annotation narrative}.
%It should be emphasized that $\bm{w}^a$ and $\bm{m}^a$ are represented in the local frame of agent $a$.
The task of the GNN is to update the intention of agents utilizing the graph structure and the node features.
Alg.~\ref{alg: modeling intention dynamics with a GNN} shows the detailed steps in updating agent intention, which follows the general procedure of a message passing GNN.
Within each major iteration, messages to each agent are constructed, followed by update of agent intention.
\figurename~\ref{fig: example graph of local traffic} delineates the two steps with schematic diagrams.

We make a few remarks on Alg.~\ref{alg: modeling intention dynamics with a GNN}.
First, recalling~\eqref{eq: unicycle dynamics for vehicles}, the motion primitive for a vehicle is 2-D, consisting of linear acceleration and angular velocity.
Therefore, it might be natural to combine the future vehicle states, $\bm{w}$, and the intention, $\bm{q}$, into a multi-channel 2-D ``image''.
The resulted 3-D tensor can be encoded with a Convolutional Neural Network (CNN) (Line~\ref{alg line: intention CNN}), which helps in reducing the overall number of parameters compared to using Multi-Layer Perceptron (MLP).
Second, data is converted to the target agent frame whenever applicable, such as Line~\ref{alg line: frame change 1} and~\ref{alg line: frame change 2}, in order to preserve the symmetry of nodes in the graph.
Finally, Alg.~\ref{alg: modeling intention dynamics with a GNN} only generates the predicted intention for the next step.
For multi-step predictions, the model could be applied recurrently using the predicted intention and sampled vehicle states as the input for the next step.

\begin{algorithm}[t]
  \DontPrintSemicolon
  \SetCommentSty{emph}

  \KwIn{Traffic Graph $\mathcal{G} = \{\mathcal{V}, \mathcal{E}\}$.}
  \KwOut{Intention of all agents $Q$.}

  \tcp{$K$ defines iterations of message passing.}
  \For{$k$ from $0$ to $K-1$}{

    \tcp{Construct the message to each node.}

    \ForEach{$(b, a)$ in $\mathcal{E}$}{
      \tcp{$T_g^a$ converts data from global frame to local frame of agent $a$.}
      $c^a \leftarrow \text{CNN}_w(T_g^a\bm{w}^a, \bm{q}^a)$\; \label{alg line: intention CNN}
      $c^b \leftarrow \text{CNN}_w(T_g^a \bm{w}^b, \bm{q}^b)$\;\label{alg line: frame change 1}
      $m_{b \rightarrow a} \leftarrow \text{MLP}_m(T_g^a \bm{x}^a, c^a, T_g^a \bm{x}^b, c^b)$\;\label{alg line: frame change 2}
    }

    \tcp{Update the intention of each node.}

    \ForEach{$a$ in $\mathcal{V}$}{
      \tcp{Element-wise max is used as Aggr.}
      $m^a \leftarrow \text{Aggr}(\{m_{b \rightarrow a}: (b, a)\in\mathcal{E}\})$\;\label{alg line: aggregation}
      $c^a \leftarrow \text{CNN}_w(T_g^a\bm{w}^a, \bm{q}^a)$\;
      $p^a \leftarrow \text{CNN}_m(T_g^a\bm{m}^a)$\;
      $\bm{q}^a \leftarrow \text{MLP}_q(T_g^a \bm{x}^a, c^a, p^a, m^a)$\;
    }
  }
  \Return{$\{\bm{q}^a : a\in\mathcal{V}\}$}

  \caption{GNN for Intention Dynamics}
  \label{alg: modeling intention dynamics with a GNN}
\end{algorithm}

\subsection{Inference conditioned on the ego motion}
\label{subsec: conditional inference}

Until this point, we have considered the traffic as an autonomous system.
However, as discussed in the introduction, it is important to infer the motion of agents conditioned on the planned motion of the ego.
The proposed GNN model can be easily adapted for this purpose.

At graph construction, we could remove the directed edges to the ego node, while keeping the intention of the ego vehicle fixed based on the planned control input.
Therefore, the ego vehicle would only broadcast its current state and intention to neighbor agents but would not be affected by their reactions.
Effectively, the modifications enable the network to perform inference conditioned on the ego motion.

As a result, $f_x$ in~\eqref{eq: unicycle dynamics for vehicles} and $f_q$ in~\eqref{eq: traffic intention dynamics} together model the controlled stochastic traffic dynamics,
\begin{equation}
  \label{eq: stochastic traffic dynamical model}
  \bm{Y}_{t+1} = f(\bm{Y}_t, \bm{u}^e_t, \bm{n}_t),
\end{equation}
which is in the same form of a general motion model as~\eqref{eq: general dynamical models}.
In~\eqref{eq: stochastic traffic dynamical model}, $\bm{u}^e$ is the control of the ego vehicle.
$\bm{n}$ is the random variable determining the controls to be executed by the agents following their intention.
Note we consider the local map $\bm{m}$ as a time-varying constant in the traffic model.
Therefore, $\bm{m}$ is not explicitly listed as inputs in~\eqref{eq: stochastic traffic dynamical model}.

\subsection{Training Loss}
\label{subsec: training loss}

The parameters in $f_q$ are learned by minimizing the cross entropy between the predicted intention $\bm{q}_t^a$ for agent $a$ and its target (ground truth) intention $\bm{q}_t^{a*}$.
Unfortunately, the ground truth intention cannot be known from a dataset.
Instead, we approximate the target intention through the known vehicle states at consecutive time steps.
One simple way is to define the target intention $\bm{q}_t^{a*}$ as a \textit{one-hot} vector, \textit{i.e.},
\begin{equation}
  \label{eq: one-hot target intention}
  \bm{q}_t^{a*}(i) =
  \begin{cases}
    1, \; \text{if } i = \underset{i}{\arg\min}\ \|\bm{x}_{t+1}^{a*} - f_x(\bm{x}_t^{a*}, \bm{u}_i)\|, \\
    0, \; \text{otherwise}.
  \end{cases}
\end{equation}
where $\bm{x}_t^{a*}$ and $\bm{x}_{t+1}^{a*}$ are the ground truth states for agent $a$ at consecutive time steps\footnote{The velocity of agents is obtained through numerical differentiation.}.
In practice, we find that it leads to overfit by defining the target intention as~\eqref{eq: one-hot target intention}  since the exact traffic state hardly shows up repeatedly.
To promote generalization, the target intention in~\eqref{eq: one-hot target intention} is ``smoothed'' to a Gaussian distribution,
\begin{equation}
  \label{eq: Gaussian target intention}
  \bm{q}_t^{a*}(i) = \frac{1}{\eta} \exp\prl{-\frac{1}{2}\|\bm{x}_{t+1}^{a*}-f_x(\bm{x}_t^{a*}, \bm{u}_i)\|^2_{\Sigma}},
\end{equation}
where $\eta$ serves as a normalization factor.
$\Sigma=\text{diag}\prl{\sigma^2_x, \sigma^2_y, \sigma^2_{\theta}, \sigma^2_v}$ represents the covariance matrix modeling the empirical uncertainty for the difference between the true and predicted vehicle states after one time step.

With the target intention defined, we could use cross entropy as the loss function shown as the following,
\begin{equation}
  \label{eq: single step loss function}
  l = \sum_{n} \sum_{a} \sum_{i} -\bm{q}_{t, n}^{a*}(i) \log(\bm{q}_{t, n}^a(i)).
\end{equation}
In~\eqref{eq: single step loss function}, $i$ indexes the elements in $\bm{q}$, $a$ indexes agents in a traffic snapshot, and $n$ indexes snapshots in the dataset.
We may also chain together consecutive snapshots, as the following, to ensure consistent intention transition over a longer period.
\begin{equation}
  \label{eq: recurrent loss function}
  l = \sum_{n} \sum_{t} \sum_{a} \sum_{i} -\bm{q}_{t, n}^{a*}(i) \log(\bm{q}_{t, n}^a(i)),
\end{equation}
where $t$ indexes different steps in a sequence, while $n$, instead of indexing snapshots, indexes snapshot sequences in the dataset.
In a sequence from the real dataset, an agent may appear or disappear at any intermediate step of the total $T$ steps, making the agent have a shorter lifetime than $T$.
Thanks to the Markovian property of RTGNN, data for agents with a shorter lifetime can still be used in~\eqref{eq: recurrent loss function} where loss of an agent is only considered when it is within the local region.

\section{Experiments}
\label{sec: experiments}

\begin{figure}[t]
  \centering
  %\subfloat[]{\includegraphics[width=0.48\textwidth]{traffic_circles_rtgnn}\label{subfig: traffic circles RTGNN}}
  %\subfloat[]{\includegraphics[width=0.48\textwidth]{traffic_circles_trajectron++}\label{subfig: traffic circles Trajectron++}} \\
  %\subfloat[]{\includegraphics[width=0.48\columnwidth]{parking_slots_rtgnn}\label{subfig: parking slots RTGNN}}
  %\subfloat[]{\includegraphics[width=0.48\columnwidth]{parking_slots_trajectron++}\label{subfig: parking slots Trajectron++}} \\
  %\subfloat[]{\includegraphics[width=0.48\columnwidth]{straight_roads_rtgnn}\label{subfig: straight roads RTGNN}}
  %\subfloat[]{\includegraphics[width=0.48\columnwidth]{straight_roads_trajectron++}\label{subfig: straight roads Trajectron++}} \\
  %\subfloat[]{\includegraphics[width=0.48\columnwidth]{intersections_rtgnn}\label{subfig: intersections RTGNN}}
  %\subfloat[]{\includegraphics[width=0.48\columnwidth]{intersections_trajectron++}\label{subfig: intersections Trajectron++}} \\
  \includegraphics[width=0.48\columnwidth]{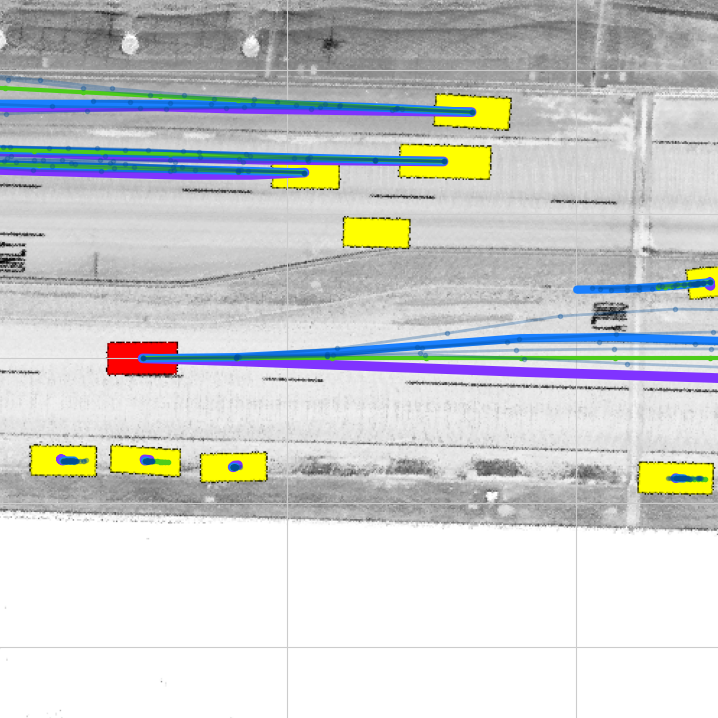}
  \includegraphics[width=0.48\columnwidth]{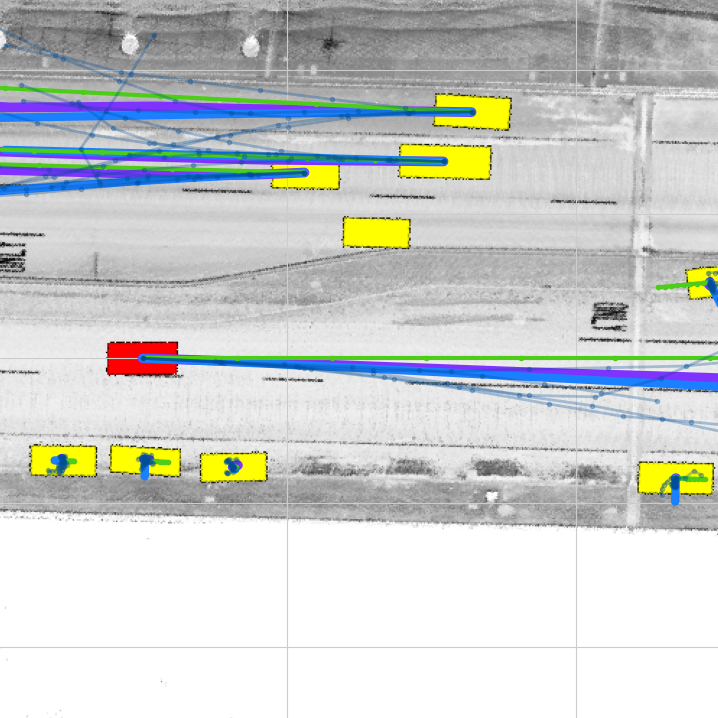} \\
  \includegraphics[width=0.48\columnwidth]{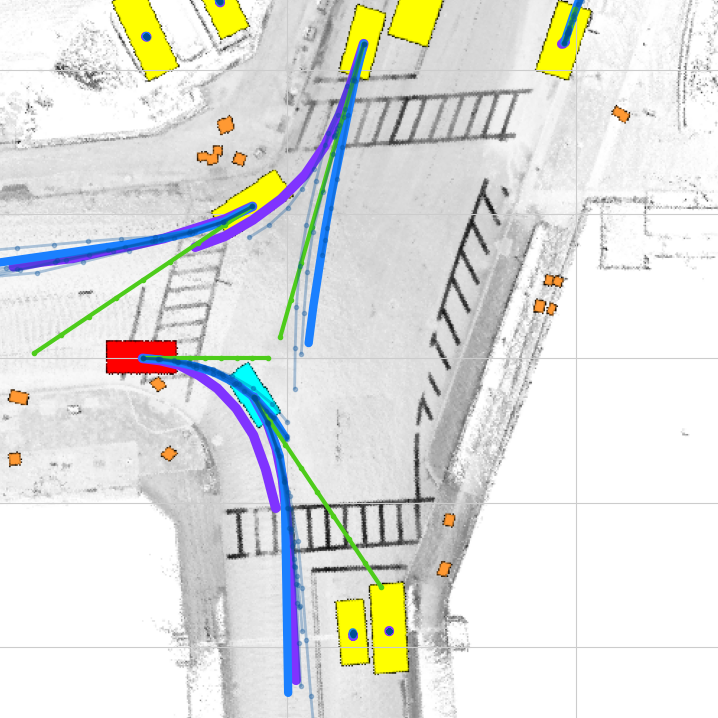}
  \includegraphics[width=0.48\columnwidth]{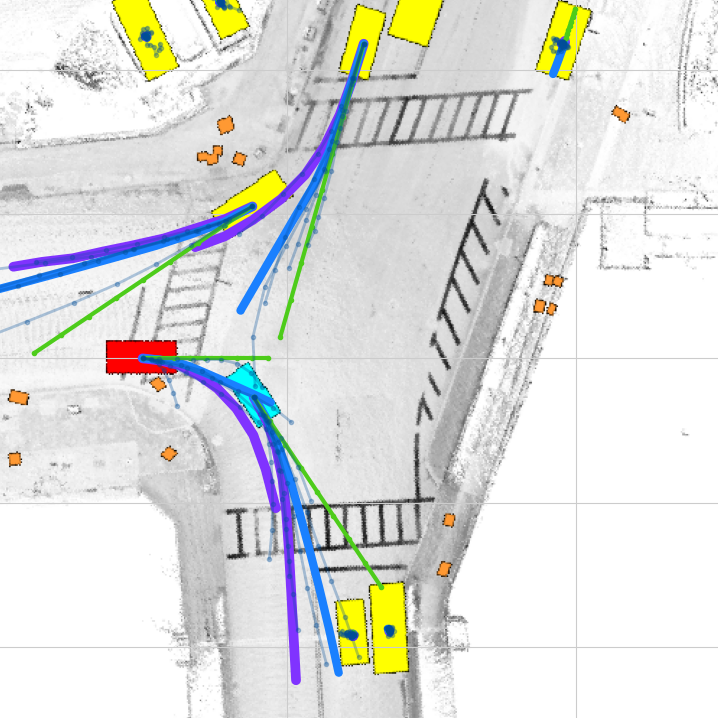} \\
  \caption{Comparisons between (left) RTGNN and (right) Trajectron++ in different scenarios, including (top) straight roads and (bottom) intersections. In the figures, the ego is shown in red, while yellow and cyan rectangles are agent vehicles. Orange squares represent pedestrians. Purple lines are the ground truth. Green lines are predictions of a constant velocity model. Blue and gray lines are the maximum likelihood and five sampled predictions.}
  \label{fig: qualitative comparison with trajectron++}
\end{figure}

In this work, we use the nuScenes dataset~\cite{Caesar-CVPR-2020}\footnote{The dataset is downloaded and used by Ke Sun, the first author of this work, when working at University of Pennsylvania.} to train and evaluate the proposed traffic model.
Specially, only the data collected at Boston Seaport\footnote{The rest of the data in nuScenes is collected in Singapore, where vehicles drive on the left side of the roads. The concern is that mixing training data of different driving conventions may have negative impact on the outcome.} is used.
Each training sample consists of four-second traffic, which includes eight time steps given the data is labeled at 2Hz.
Built upon the official split~\cite{Caesar-CVPR-2020}, the training set includes 4055 non-overlapped samples/sequences, while the validation set includes 352 samples.
Although the length of training samples is fixed, RTGNN can be applied to make predictions of arbitrary length.
The number of testing samples depends on the required prediction length.
For four-second prediction, there exists 915 non-overlapped testing samples.
We train RTGNN as an autonomous system, \textit{i.e.} treating the ego as a nominal agent vehicle.
For experiments on making predictions conditioned on ego motions, the input graph representation is modified as introduced in Sec.~\ref{subsec: conditional inference}, while the parameters for the network remain the same.
More details on the network implementation and training are provided in Appendix~\ref{sec: network setup} and~\ref{sec: training setup} respectively.

In the following, we compare the performance of RTGNN with Trajectron++~\cite{Salzmann-ECCV-2021}, a recently open-sourced work on traffic prediction.
Within the existing literature on this subject, Trajectron++ might be the most similar to this work in terms of the properties of the traffic model, which has also been evaluated on the nuScenes dataset.
Trajectron++ is able to predict trajectories for all agents in a scene jointly.
The predictions could be made conditioned on a future trajectory of the ego vehicle.
On the difference, Trajectron++ is not designed to be used as a Markovian model.
It takes historic trajectories of agents and predicts their future for multiple time steps.
Meanwhile, Trajectron++ does not explicitly model the hidden states of the agents as in this work, it adopts a CVAE framework~\cite{Sohn-NIPS-2015} which introduces discrete latent variables to encode high-level behaviors.

We make a few minor tweaks on the implementation of Trajectron++.
First, in the existing implementation of Trajectron++, future data of an agent are used to compute velocity and acceleration at the exact present time\footnote{\url{https://github.com/StanfordASL/Trajectron-plus-plus/issues/40}}.
However, the assumption is invalid in practice due to the unavailability of the future data.
Instead, we use delayed velocity and acceleration computed with only past data.
Second, parked vehicles are also included in the scene.
There are chances that parked vehicles merge into the traffic which may cause collisions with the ego if not anticipated.
With the above two modifications, we retrain Trajectron++ for twenty epochs, among which the model with the least validation error is used for testing.
In training Trajectron++, we use training samples with four seconds of future traffic data as in training RTGNN.
In testing RTGNN, one-second past data is used to compute agent velocity and acceleration at the present.
For fair comparison, past traffic data of the same length are used as input for Trajectron++.

\subsection{Qualitative Results}

\begin{figure}[t]
  \centering
  %\subfloat[]{\includegraphics[width=0.39\textwidth]{left_turn_uncond_rtgnn}
  %            \label{subfig: unconditional left turn}}
  %\subfloat[]{\includegraphics[width=0.41\textwidth]{left_turn_cond_rtgnn}
  %            \label{subfig: conditional left turn}} \\
  \includegraphics[width=0.48\columnwidth]{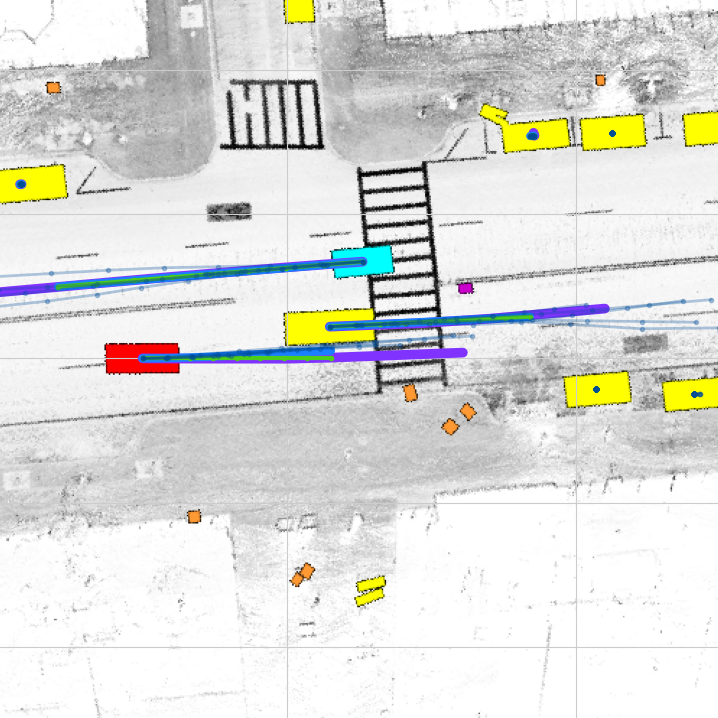}
  \includegraphics[width=0.48\columnwidth]{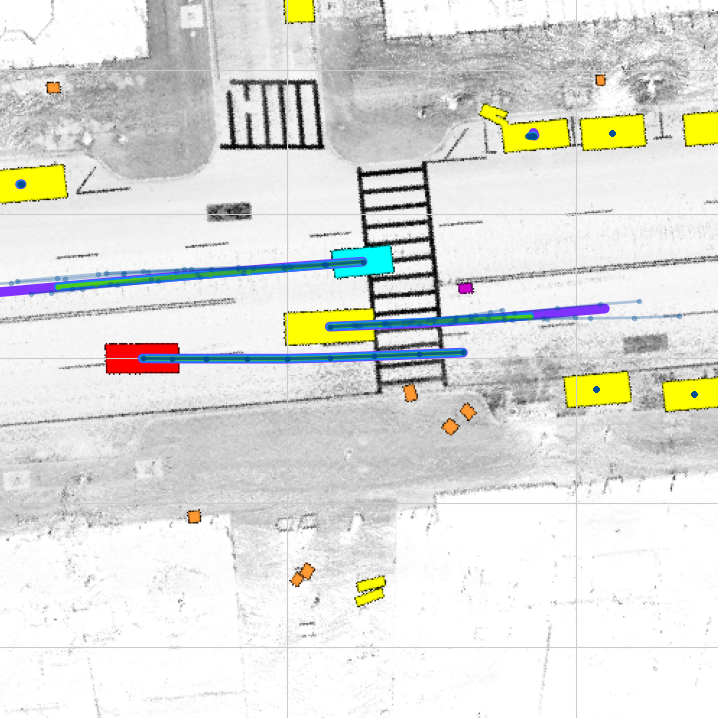}
  \caption{Comparisons between predictions (left) without and (right) with conditioning on the ego (red) motion. Given that the ego changes lanes with a high speed in the case of conditional inference (right), the leading agent vehicle defers lane change to a later time step compared to the case of unconditional inference (left). In the figures, the ego is shown in red, while yellow and cyan rectangles are agent vehicles. Orange squares represent pedestrians. Purple lines are the ground truth. Green lines are predictions of a constant velocity model. Blue and gray lines are the maximum likelihood and five sampled predictions of RTGNN.}
  \label{fig: qualitative comparison with conditional inference}
\end{figure}

\begin{table*}[t]
  \centering
  \begin{threeparttable}[t]
    \caption{Prediction accuracy of traffic models for various prediction horizon\tnote{1}}
    \label{tab: qualitative comparisons in prediction accuracy}
    \renewcommand{\arraystretch}{1.5}
    \begin{tabular}{*{9}{c|} c}
      \toprule[1.5pt]
      \multirow{2}{6em}{\centering Cases} & \multirow{2}{6em}{\centering Methods}
      & \multicolumn{2}{c|}{1s}
      & \multicolumn{2}{c|}{2s}
      & \multicolumn{2}{c|}{3s}
      & \multicolumn{2}{c}{4s} \\ \cline{3-10}
      & & M.L.P.\tnote{2} & S.P.\tnote{3}
        & M.L.P. & S.P.
        & M.L.P. & S.P.
        & M.L.P. & S.P.  \\ \midrule[1pt]
      \multirow{3}{6em}{\centering Unconditional Inference}
      & Const. Vel.  & 0.28/0.39 & -
                     & 0.57/1.05 & -
                     & 0.95/1.98 & -
                     & 1.41/3.16 & - \\ \cline{2-10}
      & Trajectron++ & \textbf{0.26/0.37} & 0.19/0.25
                     & 0.55/1.02 & 0.41/0.71
                     & 0.93/1.94 & 0.70/1.41
                     & 1.39/3.11 & 1.08/2.35 \\ \cline{2-10}
      & RTGNN        & \textbf{0.26/0.37} & \textbf{0.18/0.23}
                     & \textbf{0.50/0.89} & \textbf{0.32/0.49}
                     & \textbf{0.82/1.65} & \textbf{0.50/0.88}
                     & \textbf{1.19/2.59} & \textbf{0.70/1.32} \\ \midrule[1pt]
      \multirow{3}{6em}{\centering Conditional Inference}
      & Const. Vel.  & 0.25/0.34 & -
                     & 0.50/0.92 & -
                     & 0.83/1.74 & -
                     & 1.22/2.73 & -  \\ \cline{2-10}
      & Trajectron++ & 0.25/0.35 & 0.20/0.27
                     & 0.52/0.97 & 0.42/0.75
                     & 0.87/1.83 & 0.70/1.45
                     & 1.27/2.87 & 1.03/2.30 \\ \cline{2-10}
      & RTGNN        & \textbf{0.23/0.31} & \textbf{0.16/0.20}
                     & \textbf{0.44/0.78} & \textbf{0.28/0.43}
                     & \textbf{0.71/1.44} & \textbf{0.43/0.76}
                     & \textbf{1.01/2.20} & \textbf{0.60/1.14} \\ \bottomrule[1.5pt]
    \end{tabular}
    \begin{tablenotes}
      \item [1] There are 2714, 1505, 1111, and 915 non-overlap testing sequences for prediction horizon 1s, 2s, 3s, and 4s respectively.
      %\item [2] In unconditional and conditional inferences, traffic is considered as an autonomous system and controlled system respectively.
      %          In conditional inference, the motion of the ego vehicle is set to the corresponding ground truth.
      \item [2] M.L.P. stands for maximum likelihood predictions. The data in the corresponding entries are in the form of ADE/FDE in meters.
      \item [3] S.P. stands for sampled predictions. The data in the corresponding entries are in the form of $min_5$ADE/$\min_5$FDE in meters.
    \end{tablenotes}
  \end{threeparttable}
\end{table*}

We compare sampled and maximum likelihood predictions of Trajectron++ and RTGNN qualitatively in different scenarios, where we consider the traffic as an autonomous system.
For RTGNN, a sampled prediction is generated by sampling controls of an agent according to its intention.
The maximum likelihood predictions are produced in a similar way.
The difference is that the control with the maximum probability is selected at each time step.
Note that a trajectory obtained by sequentially selecting maximum likelihood controls may not be the overall maximum likelihood trajectory, but only serves as an approximation.
However, it would be computationally intractable to generate the true maximum likelihood predictions with the proposed traffic model in this work.

\figurename~\ref{fig: qualitative comparison with trajectron++} shows the prediction results of both methods in two traffic scenarios.
It could be observed that overall RTGNN is able to produce more sensible predictions.
Compared to Trajectron++, cases are rarer for RTGNN where the sampled or maximum likelihood predictions are infeasible, \textit{i.e.} violate the traffic rules.
Meanwhile, RTGNN generates more stable predictions for parked vehicles.
It should also be noted that, even though there is no explicit modeling of high-level behaviors, RTGNN is able to produce predictions of multi-modality, such as lane-change versus lane-keep and right-turn versus forward-move as shown in \figurename~\ref{fig: qualitative comparison with trajectron++}.

\figurename~\ref{fig: qualitative comparison with conditional inference} compares the predictions of RTGNN with or without conditioning on the ego motion in a lane change scenario.
For unconditional predictions, the traffic is considered as an autonomous system, while for conditional predictions, the ego motion is fixed to its ground truth.
In \figurename~\ref{fig: qualitative comparison with conditional inference}, different behaviors of agent vehicles can be observed depending on the ego motion.
%By fixing the ego behavior to lane change with relatively high speed, the leading agent may defer lane change to a later time step.
We would like to note that such examples are rare in the dataset.
In order to observe the difference, strong interactions should exist between the ego and agent vehicles.
Meanwhile, there should be significant difference between the predictions for the ego vehicle and the corresponding ground truth.
As a result, predictions for agents under conditional and unconditional inferences are similar in most cases.
This observation is aligned with the results presented in~\cite{Tolstaya-ArXiv-2021}.

\subsection{Quantitative Results}

We adopt commonly used metrics to quantitatively evaluate the accuracy of maximum likelihood and sampled predictions of the traffic models.
Average Displacement Error (ADE) computes the average L2 distance between the waypoints on the predicted and ground truth paths at different time steps.
Final Displacement Error (FDE), on the other hand, only computes the L2 distance of the last waypoints of the paths.
Similar to ADE and FDE, $min_k$ADE and $min_k$FDE are for sampled predictions, where only the minimum error of $k$ sampled predictions are tracked in order to avoid discouraging predictions of multi-modality.
In this work, we consider the minimum error over five sampled predictions, \textit{i.e.} $k=5$.

\tablename~\ref{tab: qualitative comparisons in prediction accuracy} compares RTGNN with Trajectron++ and a baseline constant velocity model in terms of the above metrics.
It can be observed that RTGNN is able to consistently produce more accurate maximum likelihood and sampled predictions with prediction horizons from one to four seconds.
Note compared to the results reported in~\cite{Salzmann-ECCV-2021}, the FDE of Trajectron++ in \tablename~\ref{tab: qualitative comparisons in prediction accuracy} shifts backwards in time approximately by one second.
This might be related to the fix of using past, instead of future, data to compute velocities and accelerations of the vehicles, which is mentioned at the beginning of this section.

\section{Conclusions}
\label{sec: conclusions}

In this paper, we propose a new stochastic model, RTGNN, for traffic dynamics which is able to make predictions about the future traffic state conditioned on the current  traffic state and the ego motion.
RTGNN explicitly models intention of all local agents in the traffic states and yields joint predictions of all agents of interest.
%As such, it is the first algorithm to our knowledge that can be used for motion planning and decision making.
As such, RTGNN can be seamlessly integrated into motion planning algorithms coupling prediction and decision making.
We show that RTGNN is able to generate more accurate predictions through comparisons with algorithms in previous literature.

RTGNN is the first step toward developing a stochastic traffic model that can be incorporated into a motion planning framework.
Looking ahead, we list two of many directions for future work.
First, RTGNN generates predictions based on marginal distributions of agents.
It would be desirable to draw samples from a joint distribution in order to improve the prediction consistency.
The challenge is to define a proper distribution incorporating all the agents in a scene.
Second, RTGNN currently only considers detected agents at the present.
It would be also important to model undetected agents that may appear in the future but affect the ego motion at the current moment.

\appendices
\section{Network Setup}
\label{sec: network setup}

The implementation in this work uses PyTorch\footnote{\url{https://pytorch.org/}}.
Specifically, the implementation of the GNN is based on PyTorch Geometric~\cite{Fey-ICLR-2019}\footnote{\url{https://pytorch-geometric.readthedocs.io/en/latest/}}.
%Further details about the network is provided in the following.
As introduced in Sec.~\ref{sec: modeling traffic dynamics}, we consider a discretized control space for the vehicles.
The discretized acceleration space consists of 21 accelerations equally spaced from -8m/s\textsuperscript{2} to 8m/s\textsuperscript{2}, while the discretized angular velocities consist of 21 angular velocities equally spaced from -0.5rad/s to 0.5rad/s.
The fixed duration of the motion primitives is set to 0.5s, which agrees with the labeling frequency of the nuScenes dataset.

To define the graph, we consider agents within a square of 50m around the ego vehicle.
More precisely, the boundaries of the square are set to be 40m ahead, 10m behind, and 25m on both sides of the ego vehicle.
Edges of the graph are introduced based on spatial proximity with a radius of 25m.
The number of max iterations of message passing, $K$ in Alg.~\ref{alg: modeling intention dynamics with a GNN}, is set to 2.

In Alg.~\ref{alg: modeling intention dynamics with a GNN}, \verb!CNN!\textsubscript{m} contains three layers encoding a local map of size $100\times100$ into a vector of size 32.
The three layers include 4, 8 and 16 filters with kernel size 5, 5, and 3 and stride 1.
Each convolution layer is followed by relu activation and max pooling.
\verb!CNN!\textsubscript{w} consists of two layers encoding the possible future states and intention of an agent of size $21\times 21\times 6$ to a vector of size 32.
Each layer uses 16 and 32 filters with kernel size 5 and stride 1.
In \verb!CNN!\textsubscript{w}, a convolution layer is followed by leaky relu activation and max pooling.
\verb!MLP!\textsubscript{m} consists of 3 layers encoding the message from a vector of dimension 70 to 16.
The number of hidden units in the first two layers of \verb!MLP!\textsubscript{m} are 64 and 32.
\verb!MLP!\textsubscript{q} decodes the node feature and the aggregated message from a vector of size 68 to the intention of dimension 441.
\verb!MLP!\textsubscript{q} consists of 3 layers, the first two of which have 64 and 128 hidden units.
In total, the network consists of 94,105 parameters.

\section{Training Setup}
\label{sec: training setup}

To define the target intention in~\eqref{eq: Gaussian target intention}, we set $\sigma_x=\sigma_y=5\mathrm{e}{-2}$m, $\sigma_\theta=1.75\mathrm{e}{-2}$rad and $\sigma_v=0.1$m/s.
As introduced in Sec.~\ref{sec: experiments}, each training sample accounts for four seconds of traffic, \textit{i.e.} $t=8$ in~\eqref{eq: recurrent loss function} given that data are labeled at 2Hz in nuScenes.
To train the network, we use the Adam optimizer~\cite{Kingma-ICLR-2015} implemented in PyTorch with learning rate set to $2\mathrm{e}{-3}$.
The batch size for each training iteration is 16.
We set the maximum epochs to be 50.
It takes about 10 hours to train on a desktop with an Intel i9-9920X CPU (12 cores at 3.5GHz), 32G RAM, and a Nvidia 2080Ti GPU.
With the same hardware configuration, it takes 0.176s on average to make one four-second sample prediction.
The time required varies depending on the number of agents in the scene.

To alleviate the compounding error issue of recurrent neural networks, we apply scheduled sampling as in~\cite{Bengio-NIPS-2015}.
Specifically, the sampling rate (the rate of using sampled vehicle states instead of ground truth states as the input to the next step) increases linearly from 0.0 at Epoch 10 to 0.5 at Epoch 30.
The sampling rate is constant otherwise.

\section*{Acknowledgment}
The authors gratefully acknowledge the support of Qualcomm Research who sponsored this work.
We would also like to thank Dr. Pratik Chaudhari from University of Pennsylvania, Boris Ivanovic and Dr. Marco Pavone from Stanford University for their suggestions in improving this work.

\addtolength{\textheight}{-5cm}

\bibliographystyle{IEEEtran}
\bibliography{IEEEabrv,ref}

\end{document}